\documentclass[10pt,twocolumn,letterpaper]{article}

\usepackage{iccv}
\usepackage{times}
\usepackage{epsfig}
\usepackage{graphicx}
\usepackage{amsmath}
\usepackage{amssymb}

\usepackage{multirow}
\usepackage{booktabs}
\usepackage{threeparttable}
\usepackage{makecell}

\usepackage{caption}
\usepackage{lipsum}

\usepackage[pagebackref=true,breaklinks=true,letterpaper=true,colorlinks,bookmarks=false]{hyperref}

\usepackage[capitalize]{cleveref}
\crefname{section}{Sec.}{Secs.}
\Crefname{section}{Section}{Sections}
\Crefname{table}{Table}{Tables}
\iccvfinalcopy % *** Uncomment this line for the final submission

\def\iccvPaperID{2018} % *** Enter the ICCV Paper ID here

\begin{document}

%%%%%%%%% TITLE
\title{ FastMESH: Fast Surface Reconstruction by Hexagonal Mesh-based\\ Neural Rendering }

\author{Yisu Zhang$^1$, \ \ \ Jianke Zhu$^{1,2}\thanks{Corresponding author is Jianke Zhu.}$\ \ \ Lixiang Lin$^1$, \\
 $^1$Zhejiang University \\
 $^2$Alibaba-Zhejiang University Joint Research Institute of Frontier Technologies \\ 
 {\tt\small \{zhyisu, jkzhu, lxlin\}@zju.edu.cn}
}

\twocolumn[{%
\renewcommand\twocolumn[1][]{#1}%
\maketitle
\vspace{-2.8em}
\begin{center}
    \centering
    \includegraphics[trim={12.5cm 13.5cm 3.5cm 1cm},clip,width=1.0\linewidth]{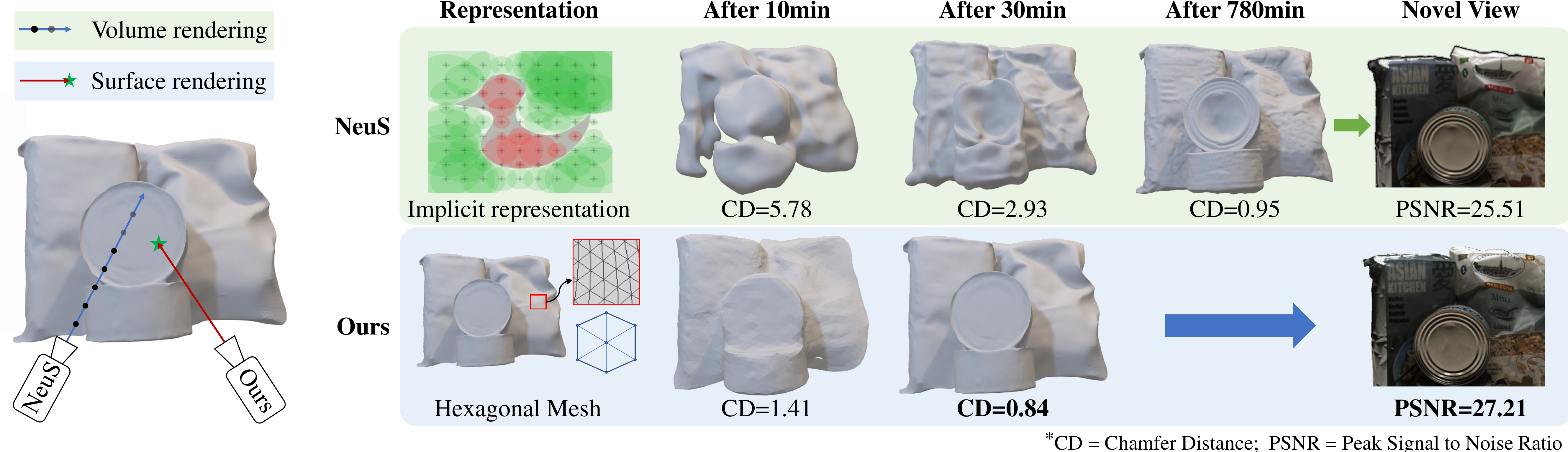}
    % \vspace{-0.2cm}
    \captionof{figure}{\textbf{Comparison with NeuS~\cite{Wang:2021:Neus}.} In contrast to the neural implicit method, our proposed FastMesh approach employs an explicit hexagonal mesh, which renders the surface solely based on the intersection point rather than sampling ray in volume rendering. Such a scheme effectively disentangles the geometry and appearance so that our method achieves 20-fold acceleration compared to NeuS while obtaining more accurate results on both reconstruction and novel view synthesis.}
    % CD = Chamfer Distance, PSNR = Peak Signal to Noise Ratio
    % \vspace{-0.2cm}
    \label{fig:intro_vis}
\end{center}
}]

% \maketitle
\ificcvfinal\thispagestyle{empty}\fi

%%%%%%%%% ABSTRACT
\begin{abstract}
\vspace{-10mm}
   
   Despite the promising results of multi-view reconstruction, the recent neural rendering-based methods, such as implicit surface rendering (IDR) and volume rendering (NeuS), not only incur a heavy computational burden on training but also have the difficulties in disentangling the geometric and appearance. Although having achieved faster training speed than implicit representation and hash coding, the explicit voxel-based method obtains the inferior results on recovering surface. To address these challenges, we propose an effective mesh-based neural rendering approach, named FastMESH, which only samples at the intersection of ray and mesh. A coarse-to-fine scheme is introduced to efficiently extract the initial mesh by space carving. More importantly, we suggest a hexagonal mesh model to preserve surface regularity by constraining the second-order derivatives of vertices, where only low level of positional encoding is engaged for neural rendering. The experiments demonstrate that our approach achieves the state-of-the-art results on both reconstruction and novel view synthesis. Besides, we obtain 10-fold acceleration on training comparing to the implicit representation-based methods. 

\end{abstract}

\begin{figure*}
  \begin{center}
     \includegraphics[trim={0cm 0cm 0cm 0cm},clip,width=1.0\linewidth]{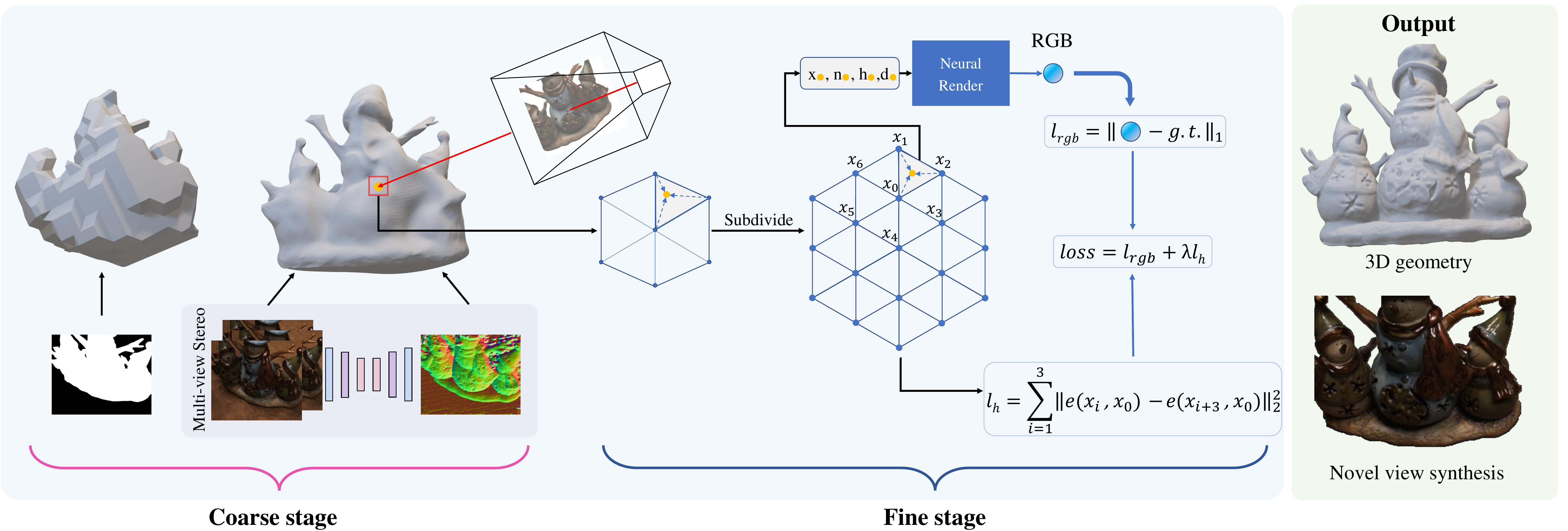}
  \end{center}
  \vspace{-0.4cm}
  \caption{{\bf Overview of our proposed two-stage coarse-to-fine pipeline.} The initial mesh is obtained by space carving on masks, which is further refined by normals from multi-view stereo~\cite{Gu:2020:cascade}. At the fine stage, the coarse hexagonal mesh is subdivided to capture more detailed geometry. The interpolated intersection attributes are fed into neural renderer to predict the RGB value for each pixel, where the mesh regularization is imposed to preserve surface regularity during optimization.} 
  \label{fig:framework}
  \vspace{-0.5cm}
\end{figure*}

%%%%%%%%% BODY TEXT
\section{Introduction}
Recovering scene geometry and appearance from multi-view images is a fundamental problem in computer vision, which has a large number of real-world applications.

Recently, the neural rendering-based methods, such as implicit surface representation~\cite{Yariv:2020:multiview, Niemeyer:2020:DVR, sitzmann:2019:scene} and neural volume rendering~\cite{Wang:2021:Neus, Yariv:2021:volsdf, Mueller:2022:instantngp, Oechsle:2021:UNISURF, Mildenhall:2020:NeRF, Sun:2022:Direct}, have achieved the impressive progress in both geometric reconstruction and novel view synthesis. The surface-based methods represent the continuous implicit surface as a level set, which renders the appearance from surfaces. Alternatively, the volume rendering methods integrate the densities by drawing samples along the rays. The key to these approaches is to train a deep multi-layer perceptron (MLP) that predicts the appearance and geometry information from the sampled points along each view ray. However, a single fully-connected MLP may lead to oversmoothing geometry and low-quality rendered images. To deal with the highly coupled geometry and appearance, it usually takes several hours to train an MLP. This greatly hinders them from some application scenarios with limited computational power.

To facilitate the fast training in NeRF, some approaches~\cite{Mueller:2022:instantngp, Radu:2022:HashSDF} employ the hash-based positional encoding. Despite the state-of-the-art results on novel view synthesis, their recovered scene geometry may deviate severally from the ground-truth. Instead of using the signed distance (SDF) or occupancy field, the explicit volumetric representations~\cite{Sun:2022:Direct, Wu:2022:Voxurf} are employed to accelerate volume rendering, which directly optimizes the geometric and color information of explicit voxel grids through interpolation. However, they may not reconstruct the accurate surface to represent the precise geometry of the object or scene. Additionally, the point cloud-based methods~\cite{fan:2017:point, lin:2018:learning, mandikal:2018:3d} have some outliers in their reconstruction results while the synthesized novel views cannot reflect the real lighting conditions. Mesh is a kind of simple explicit 3D representation that captures the geometric details with topology.  Without neural rendering, the previous mesh-based methods~\cite{Goel2021Differentiable, Worchel:2022:nds} obtain inferior results on both reconstruction and novel view synthesis through a differentiable render only.%  the geometry and synthesis results are not satisfactory. 

To address the above limitations, we propose an effective mesh-based neural rendering approach to fast surface reconstruction and novel view synthesis. It has the merit of only sampling at the intersection of ray and mesh rather than multiple points along the ray in volume rendering-based method, which enables to accurately disentangle of geometry and appearance from input images. Moreover, an efficient two-stage coarse-to-fine scheme is introduced so that the initial mesh can be effectively extracted by space carving~\cite{Laurentini:1994:VisualHull}. Furthermore, we employ an isotropic hexagonal mesh that not only preserves the structure regularity by constraining the second-order derivatives of mesh vertices but also ensures fast subdivision. By taking advantage of the presented mesh-based scheme, only few level of positional encoding is engaged for neural rendering, which substantially accelerates the training process.

In summary, our main contributions are: 1) an effective coarse-to-fine scheme for multi-view surface reconstruction based on explicit mesh-based neural rendering; 2) the hexagonal regularization for stable neural surface reconstruction and fast subdivision; 3) in contrast to the state-of-the-art methods, our proposed approach achieves the best reconstruction and novel view synthesis performance while having 10-fold acceleration in training.

%-------------------------------------------------------------------------
\section{Related Work}
During the past decades, the extensive research efforts have been devoted to multi-view reconstruction. We briefly review the related work on classical multi-view stereo algorithms, neural implicit surface reconstruction, and explicit surface-based methods.

\subsection{Classical Multi-View Stereo}

The conventional multi-view stereo (MVS) methods make use of various 3D representations, such as mesh~\cite{Fua:1995:Object, Furukawa:2006:CarvedVisualHull}, point cloud~\cite{Furukawa:2009:accurate, Lhuillier:2005:quasi}, voxel~\cite{Kutulakos:2000:theory, Seitz:1999:photorealistic} and depth map~\cite{Campbell:2008:using, Galliani:2015:Massively, Schonberger:2016:pixelwise}. These approaches usually have a complex pipeline and require post-processing operations like deep fusion. The results of traditional MVS may have obvious artifacts in the cases of cluttered scenes, large matching noise, and poor correspondences.

Recently, the learning-based MVS approaches are proposed to robustly estimate depth maps. For instance, 3D cost volume~\cite{Yao:2018:Mvsnet, Gu:2020:cascade, Luo:2019pmvs, Chen:2020:TPAMI, Zhang:2020:visibility} is built to aggregate the warped features. Then, the depth map is regressed through a 3D CNN. Moreover,~\cite{zagoruyko:2015:learning, luo:2016:efficient, sitzmann:2019:scene} learn to match 2D features. Regardless of conventional MVS or learning-based approaches, most of them employ point clouds as 3D representation. To obtain the explicit mesh, they require an additional triangulation step like Poisson surface reconstruction~\cite{Kazhdan:2006:poisson}. Although having restored the accurate geometric shapes with details by such representation, it may lead to outliers and poor results in synthesizing the novel views.

\subsection{Neural Implicit Surface Representation}
With the prevalence of neural rendering work~\cite{Mildenhall:2020:NeRF, Aliev2020npbg, Tewari:2021:AdvancesNeuralRendering, Sun:2022:Direct, Mueller:2022:instantngp}, neural implicit functions have recently emerged as representation of geometry ~\cite{michalkiewicz:2019:implicit, genova:2019:learning, Park:2019:DeepSDF, xu:2019:disn, saito:2019:pifu, peng:2020:convolutional, Mescheder:2019:OccupancyNetworks, atzmon:2019:controlling} and appearance ~\cite{liu:2020:NeuralSparse, liu:2020:dist, oechsle:2019:texture, schwarz2020graf, sitzmann:2019:scene}. They map 3D coordinates into scalar values that represent the probability of the point lying inside the object or the distance from point to surface. As most of these methods need 3D ground truth for supervision, several studies~\cite{Liu:2019:LearningImplicit, Niemeyer:2020:DVR, Yariv:2020:multiview, Wang:2021:Neus, toussaint:2022:fast, zhang:2021:MVSDF, fu:2022:Geo, Yariv:2021:volsdf, darmon:2022:improving} combine neural implicit function with differentiable rendering to restore scenes directly from 2D images in a self-supervised manner. 

Jiang et al.~\cite{jiang:2020:sdfdiff} define a differentiable shading function, which calculates the signed distance at any position by interpolating dense grids. Neural implicit surface rendering~\cite{Yariv:2020:multiview, Niemeyer:2020:DVR} is used to obtain the appearance, where the differentiable rendering formulation using implicit gradients of surfaces is engaged to recover high-frequency details of geometry and color. However, a good initialization is usually required. Moreover, it takes several hours for training and performs inferior to the volume rendering-based method on novel view synthesis.

NeRF~\cite{Mildenhall:2020:NeRF} and its variants~\cite{Kaizhang:2020:NeRF++, Yu:2020:PixelNerf, barron:2021:mip, niemeyer:2021:giraffe} make use of volume rendering to synthesize radiation fields along rays by learning MLPs, which achieve the stunning results on novel view synthesis.  UNISURF~\cite{Oechsle:2021:UNISURF} optimizes an occupancy function, and others~\cite{Yariv:2021:volsdf, Wang:2021:Neus} extend the signed distance field (SDF) through volumetric rendering. Although volume rendering-based approach does not require the labeled mask for supervision, it is still computationally intensive to reconstruct the detailed geometry and appearance.

\subsection{Explicit Surface Representation}
%Neural implicit representation becomes a promising approach to recovering 3D shape from multi-view images. 

Besides neural implicit surface, the explicit representations, such as point cloud, voxel and mesh, are also explored in recovering 3D shape from multi-view images. 

Voxel grids~\cite{fridovich:2022:plenoxels, chen:2022:tensorf, Sun:2022:Direct, Wu:2022:Voxurf} are widely used to represent 3D shapes by combining with volume rendering to encode geometry and appearance information into grids. However, the major drawback is that the voxel resolution limits their capability of faithfully recovering scene geometry.

In contrast, mesh is the more compact and flexible representation for 3D shapes. Some recent methods~\cite{Goel2021Differentiable, Worchel:2022:nds} directly optimize the geometry and appearance of mesh to reconstruct 3D objects. Instead of using discrete operation, a continuous function is used to render mesh onto image by rasterization~\cite{nimier:2020:radiative, cole:2021:differentiable, bangaru:2020:unbiased, Laine:2020:diffrast, liu:2019:softras, kato:2018:neural}, which enables back-propagation to facilitate neural rendering.

To this end, some studies~\cite{Li:2018:Redner, NimierDavid:2019:Mitsuba2, Zhang:2020:PathDiffRend} make use of differentiable renderer to jointly optimize the geometry and albedo. Due to the priors like lighting conditions, it is hard for them to generalize for the arbitrary scene. Although mesh representation offers a faster processing speed than others, the final geometric details and novel view synthesis results may not be satisfactory due to the lack of reasonable constraints~\cite{Nicolet:2021:LargeSteps}.

To deal with the problem that a large number of ray samples traverse the empty space, NSVF~\cite{liu:2020:NeuralSparse} employs the octree data structure to represent voxels. Such structure accelerates the transversal while requiring pruning and updating the octree. Sun et al.~\cite{Sun:2022:Direct} represent the model as local features from grids, which is decoded by a hybrid architecture including voxel and shallow MLP. Instant-NGP~\cite{Mueller:2022:instantngp} propose multi-resolution hash encoding to avoid the complex pruning and updating operations in the octree. It significantly reduces the training time and achieves encouraging novel view synthesis. However, the recovered geometric details are quite poor.

\section{Methodology}
Given a set of images $\mathcal{I}=\left\{I_1, \cdots, I_n\right\}$ captured by calibrated cameras and their corresponding masks $\mathcal{M}=\left\{M_1, \cdots, M_n\right\}$, our objective is to estimate the 3D surface $\mathcal{G}$ of the depicted objects from the images $\mathcal{I}$ and synthesize the photo-realistic novel views. In this paper, the surface is represented by a triangulated mesh $\mathcal{G}=(V, \mathcal{F}, f)$, consisting of vertex positions $V$, triangular facet $\mathcal{F}$ and feature vector $f$ bound to vertices.

\cref{fig:framework} illustrates the whole pipeline of our method. Firstly, we design a two-stage procedure that achieves a coherent coarse shape and subsequently restores the fine geometric details. Secondly, we suggest a mesh-based neural rendering approach that reduces the multiple sampled points in volume rendering into a single point at the intersection of a ray and mesh. This enables to disentangle geometry and appearance from image color so that we can recover accurate surface geometry and synthesize novel views. Finally, we employ the hexagonal mesh that takes advantage of efficient and stable internal force regularization through enforcing the second-order derivatives of mesh vertices. It ensures the faithful reflection of the geometric shape while maintaining the topological structure.

\subsection{Coarse Initialization}
Due to the inherent difficulty in altering mesh topology during the optimization process, it becomes imperative to obtain the coarse geometry as initialization. In this paper, we resort to a mask-based approach, such as space carving~\cite{Laurentini:1994:VisualHull}. Decoupling of geometry and appearance is the cardinal prerequisite to achieve expeditious convergence of the model. Subsequently, the low mesh resolution can be viewed as a rough estimation of geometry, which approximates correctness and effectively obviates the likelihood of entrapment within local optima. To this end, we employ the normal maps estimated by multi-view stereo method~\cite{Gu:2020:cascade}  as supervision in optimization, which enables the rapid convergence of mesh geometry at the early stage. 

While the normals estimated by multi-view stereo technique may contain numerous errors and inconsistencies across multi-angle views, they nevertheless remain qualified as the coarse initialization. Thus, a rough low-resolution mesh can be procured with minimal optimization iterations. To accurately and faithfully capture the geometric intricacies sans texture, the coarse mesh is further undergone subdivision and refinement in the later optimization process.

\subsection{Mesh-based Neural Rendering}
% \subsection{ Fine Optimization}

The majority of neural implicit representation-based methods make use of volume rendering to acquire pixel intensities. The ray emitted from a pixel is denoted as $\{\mathbf{p}(t)=\mathbf{o}+t \mathbf{v} \mid t \geq 0\}$, where $o$ is the center of the camera and $v$ is the direction vector of the ray. For volume rendering, the color is approximated by accumulating along the ray by
\begin{equation}
\hat{C}=\sum_{i=1}^n T_i \alpha_i c_i,  T_i=\prod_{j=1}^{i-1}\left(1-\alpha_j\right)
\end{equation}

where $C$ is the output color, $\alpha_i$ is a discrete opacity values, and $T_i$ the accumulated transmittance. Various weighting functions $\alpha_i$ are employed by different methods~\cite{Mildenhall:2020:NeRF, Wang:2021:Neus, Yariv:2021:volsdf}. Among them, the unbiased estimation-based method in NeuS~\cite{Wang:2021:Neus} demonstrates the encouraging performance.

Instead of using the implicit representation, we propose a neural mesh rendering approach in this paper. It is akin to volume rendering, whereby the final color of a given pixel is determined by the attributes of intersection point between the ray and surface of the object, as well as the viewing direction. Accordingly, we reformulate the integral process described in Equation ($1$) as follows
\begin{equation}
\hat{C}(\mathbf{r})=\sum_{i=1}^N t\left(\mathbf{x}_i\right) \prod_{j<i}\left(1-t\left(\mathbf{x}_j\right)\right) c_\theta
\end{equation}
$t(x_i) \in \{0,1\}$ indicates whether $x_i$ is on the mesh surface. It takes $t = 0$ in free space and $t = 1$ when $x_i$ is the intersection. As in~\cite{Yariv:2020:multiview}, we condition the color field $c_{\theta}$ on points position $\mathbf{x}_i$, feature vector $\mathbf{h}_i$, surface normal $\mathbf{n}_i$ and view direction $\mathbf{d}$. 

\begin{figure}
  \begin{center}
     \includegraphics[trim={0.1cm 2cm 0.1cm 0.1cm},clip,width=1.0\linewidth]{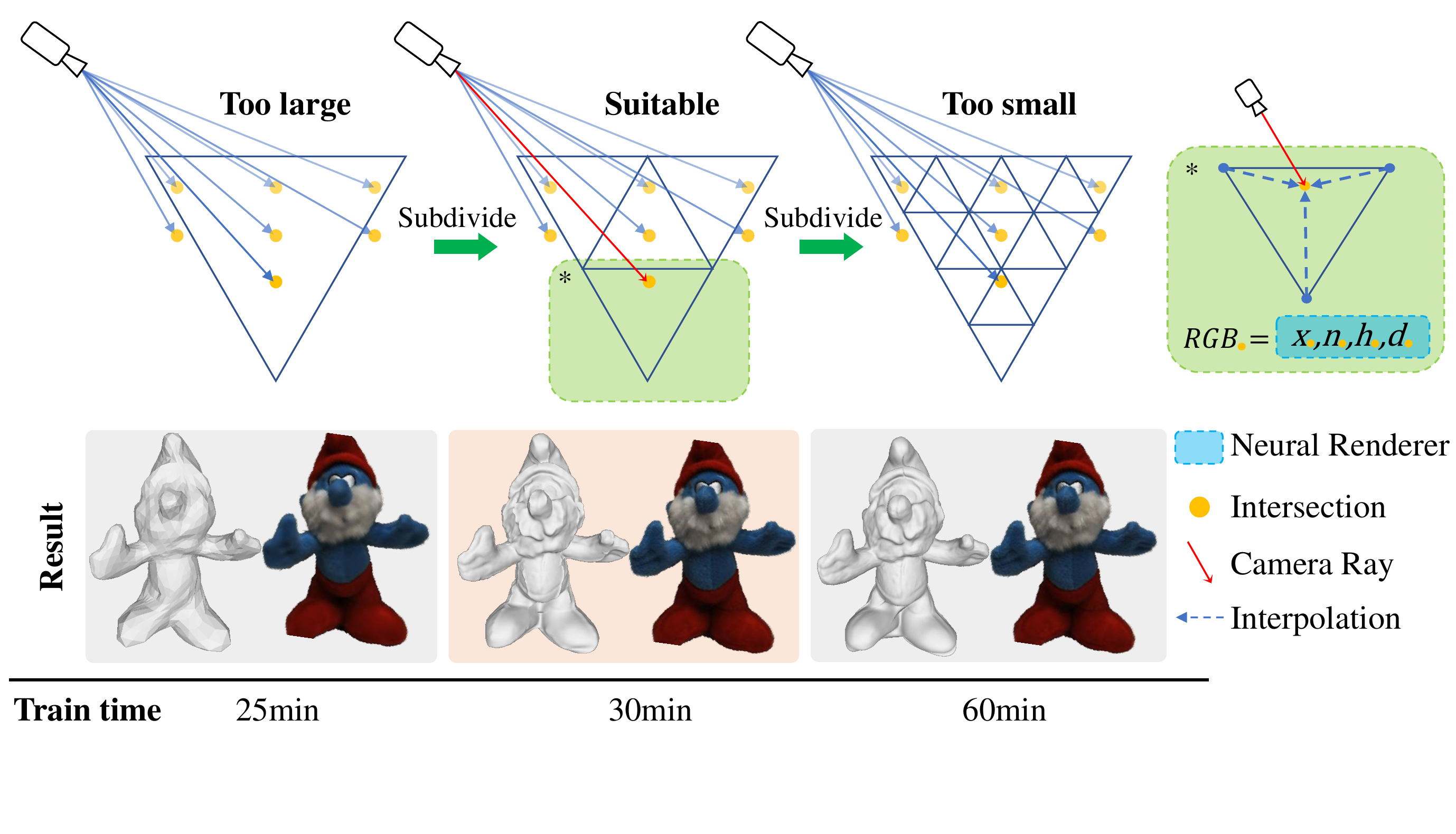}
  \end{center}
  \vspace{-0.5cm}
  \caption{{\bf Mesh-based Neural rendering. } The adequate mesh resolution is crucial to accurately capture geometric details while minimizing the computational burden associated with vertex attribute calculations.
  } 
  \label{fig:neural_render}
  \vspace{-0.5cm}
\end{figure}

Mesh is typically represented by triangular facets so that the intersection points between rays and meshes are located within these facets. The positional and feature vector information that describe meshes are bound to their vertices. Thus, the attributes of intersection points are calculated by interpolating the three vertices of triangular facet to which the intersection point belongs, as illustrated in~\cref{fig:neural_render}. 
The efficient differentiable rendering technique~\cite{Laine:2020:diffrast} ensures the feasibility of neural mesh rendering.

Therefore, the formulation of mesh-based neural rendering is simplified as below
\begin{equation}
\hat{C}(\mathbf{r})=c_\theta\left(\mathbf{x}_m, \mathbf{n}_m, \mathbf{h}_m, \mathbf{d}\right)
\end{equation}
where $\mathbf{x}_m$ denotes the position, and $\mathbf{n}_m$ is the normal. $\mathbf{h}_m$ represents the feature vector of the intersection point between the ray and the surface. $\mathbf{d}$ is the direction of ray. As shown in \cref{fig:neural_render},  $\mathbf{x}_m$, $\mathbf{n}_m$ and $\mathbf{h}_m$ are obtained through interpolation. It is necessary to  match the density of rays with the resolution of the mesh in order to achieve an optimal intersection between rays and mesh geometry. This requires a delicate balance, as excessively large triangles may yield multiple intersections with the same triangle. On the other hand, certain regions in mesh may not be traversed by rays due to excessively small triangles, which incurs the extra computations.

A fundamental assumption underlying the effectiveness of the neural mesh rendering equation is that the object being rendered is both solid and opaque. Therefore, the final color of a pixel is solely determined by the attributes of the first intersection point between the ray and mesh.

\subsection{Hexagonal Mesh Regularization}
\begin{figure}
  \begin{center}
     \includegraphics[trim={0.1cm 0.1cm 0.1cm 0.1cm},clip,width=1.0\linewidth]{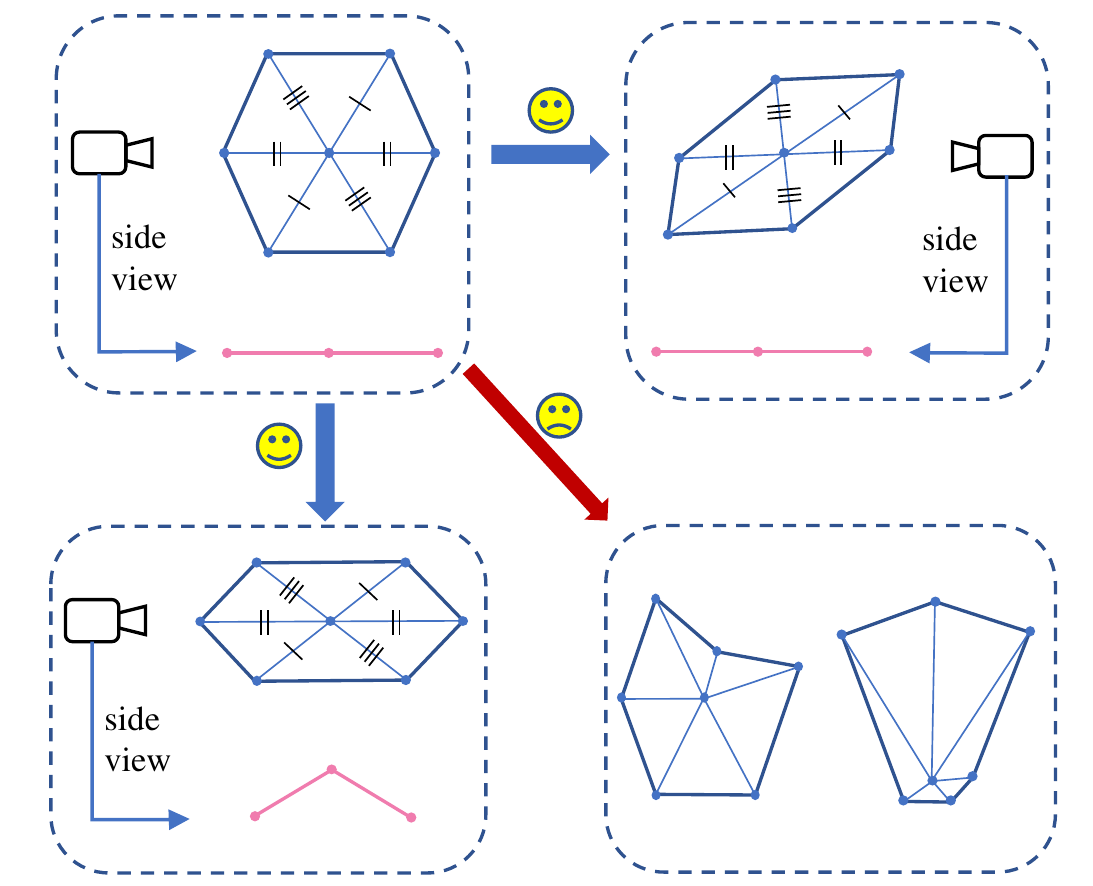}
  \end{center}
  \vspace{-0.5cm}
  \caption{{\bf Mechanism of hexagonal mesh regularization.} The hexagonal mesh regularization allows for symmetric deformations such as those in upper right and bottom left and penalizes the unstable deformation in bottom right. } 
  \label{fig:regularization}
  \vspace{-0.5cm}
\end{figure}

With a differentiable renderer, a common issue is that the predicted triangular mesh cannot be utilized due to gradient steps pulling on the silhouette without considering distortions or self-intersections. To address this problem, the regularization is often employed to enforce the structure regularity of the mesh. However, the existing mesh regularization techniques~\cite{Worchel:2022:nds, Luan:2021:Unified}, such as Laplace smoothing, cannot effectively trade-off between smoothness and the detailed geometric structures. To tackle this challenge, we take advantage of the isotropic hexagons~\cite{Fua:1995:Object} as the basic unit for mesh representation, where each unit is composed of six triangular faces, as shown in~\cref{fig:regularization}. The neighbor edge is defined as below
$$
% neighbor(e(v_j, v_i)) = e(v_{j+3}, v_i)
\mathcal{N}(e(v_j, v_i)) = e(v_{j+3}, v_i)
$$
where $\mathcal{N}(e)$ denotes the neighbor of edge $e$.  $v_i$, $v_j$ denote the vertices of hexagon.

The regularization term for each hexagonal cell ensures that the inter-vertex distances are equal. This condition implies that the coordinates of a vertex $v_i$ are equal to the average of the coordinates of its neighboring vertices $v_j$ and $v_{j+3}$, for any vertex $j$:
\begin{equation}
\mathcal{L}_h(\mathcal{S})=\sum_{i=1}^{n_v} \sum_{\substack{j=1 \\ k=N_i(j) \\ k^{\prime}=N_i(j+3)}}^3\left(2 x_i-x_k-x_{k^{\prime}}\right)^2
\end{equation}
where $\mathcal{S}$ represents vertices of mesh. $x$ denotes the coordinate of vertex along the $x$-axis. The same principle is applied to the $y$-axis and $z$-axis coordinates. Let $\mathcal{L}_h$ denote the regularization term based on hexagonal mesh. Equation $(4)$ can be rewritten into matrix form as follows
\begin{equation}
\mathcal{L}_h(\mathcal{S})=\mathbf{S}^{\top} \mathcal{K} \mathbf{S}
\end{equation}
\begin{equation}
\mathcal{K}=\left[\begin{array}{ccc}
K &   &   \\
  & K &   \\
  &   & K
\end{array}\right]
\end{equation}
where $\mathbf{S}$ is a one-dimensional vector concatenated by three-dimensional coordinates of vertices. The sparse and banded matrix $K$ is determined by the underlying structure of explicit mesh model~\cite{Fua:1995:Object}. The constraint enforces the equality of sides in hexagons, which can be mathematically formulated as a condition. It guarantees that every hexagon possesses centrosymmetry, whereby the geometric shape remains unchanged after rotating 180 degrees around its center point.

The above regularization term ensures the stability during the optimization with large steps, while also enabling the precise capture of geometric details through the use of deformations. Moreover, the isotropic remeshing is used to guarantee the attainment of high-precision details in mesh, which maintains hexagons as the fundamental unit of mesh. Thereby, the hexagonal regularization is capable of circumventing issues such as self-intersection and significant variations in triangle. Consequently, we can recover the finer geometric details while optimizing at larger steps. Thus, the accurate mesh can be obtained.

\subsection{Optimization}
In this paper, we suggest a coarse-to-fine scheme to achieve the stable optimization in the case of large steps. To alter topology and increase resolution, the conventional mesh-based methods~\cite{Worchel:2022:nds, yu:2013:reconstructing} typically employ the EL Topo tool~\cite{Brochu:2009:ElTopo} to facilitate remeshing operations during optimization. Alternatively, we adopt the fast and convenient interpolation-based subdivision operation by adding a point at the midpoint of each edge to strike a balance between speed and geometric shape. This subdivision produces four times the number of points and faces comparing to the original mesh while maintaining the hexagonal topology. To accurately recover the surface geometry, we train the first 100 iterations through a curvature loss as follows

\begin{equation}
\mathcal{L}=\mathcal{L}_{r g b} + \lambda_1 \mathcal{L}_{h} + \lambda_2 (\mathcal{L}_{\text {mask }} + \mathcal{L}_{\text {normal }})
\end{equation}
where $\mathcal{L}$ is total loss. $\mathcal{L}_{r g b}$ is the photometric loss between rendered images and reference images. $\mathcal{L}_{h}$ is the mesh regularization in Equation $(5)$. $\mathcal{L}_{\text {mask}}$ and $\mathcal{L}_{\text {normal }}$ are mask loss and normal loss, respectively. Once the approximately accurate geometric mesh and color network are obtained, we no longer rely on the supervision of error-prone normal information from multi-view stereo method. Instead, our goal shifts towards the restoration of fine-grained textures through the image-based information:
\begin{equation}
\mathcal{L}=\mathcal{L}_{r g b} + \lambda_1 ^\prime \mathcal{L}_{h} + \lambda_2 \mathcal{L}_{\text {mask }}
\end{equation}
where the $\lambda_1^\prime$ and $\lambda_3$ are weight of different regularization terms. $\mathcal{L}_{r g b}$, $\mathcal{L}_{\text {mask }}$ are denote as
\begin{equation}
\mathcal{L}_{r g b} = \sum_{\mathbf{r} \in \hat{\mathcal{M}}}\left\|\hat{C}_m(\mathbf{r})-C(\mathbf{r})\right\|_1
\end{equation}
\begin{equation}
\mathcal{L}_{\text {mask }} = 1 - \mathbf{IOU}(\hat{\mathcal{M}}, \mathcal{M})
\end{equation}
Subsequently, we perform remeshing operations at the 100th, 200th and 400th iterations, respectively. As the mesh resolution increases, the implicit reduction in the size of hexagonal cells weakens the strength of this loss term. The weakened smooth constraint and higher resolution are used to describe the geometry more accurately, facilitating the recovery of geometric details.

\section{Experiments}
\subsection{Implementation Details and Setup}
In all experiments, we fit the mesh model to multi-view images of single-scene. We employ PyTorch and Adam optimizer with an initial learning rate of $lr=1e^{-3}$, which is decayed at the 100th, 200th, and 400th iterations. To ensure the capability of differentiable rasterization, we make use of the off-the-shelf renderer nvidiffrast~\cite{Laine:2020:diffrast} in our pipeline. Moreover, a neural network similar to~\cite{Aliev2020npbg} was employed as the rendering network. The input of neural renderer consists of four components, including the intersection position $\mathbf{x}$, the normal vector $\mathbf{n}$, the feature vector $\mathbf{h}$, and the viewing direction $\mathbf{d}$. They are embedded into 3D coordinates via positional encoding that is further fed into the render network. We empirically found that the best performance is achieved in case of $k=3$ for 3D location $\mathbf{x}$. The individual objective terms are weighted with $\lambda_1 = 2$, $\lambda_2 = 50$ and $\lambda_1^\prime = 4$. To ensure the consistency and fairness of our experiments, we employed NVIDIA RTX 2080Ti GPU with 11GB GPU memory in both training and testing for all methods.

\begin{figure*}[h]
  \begin{center}
     \includegraphics[trim={0.1cm 0.3cm 3cm 1.1cm},clip,width=1.0\linewidth]{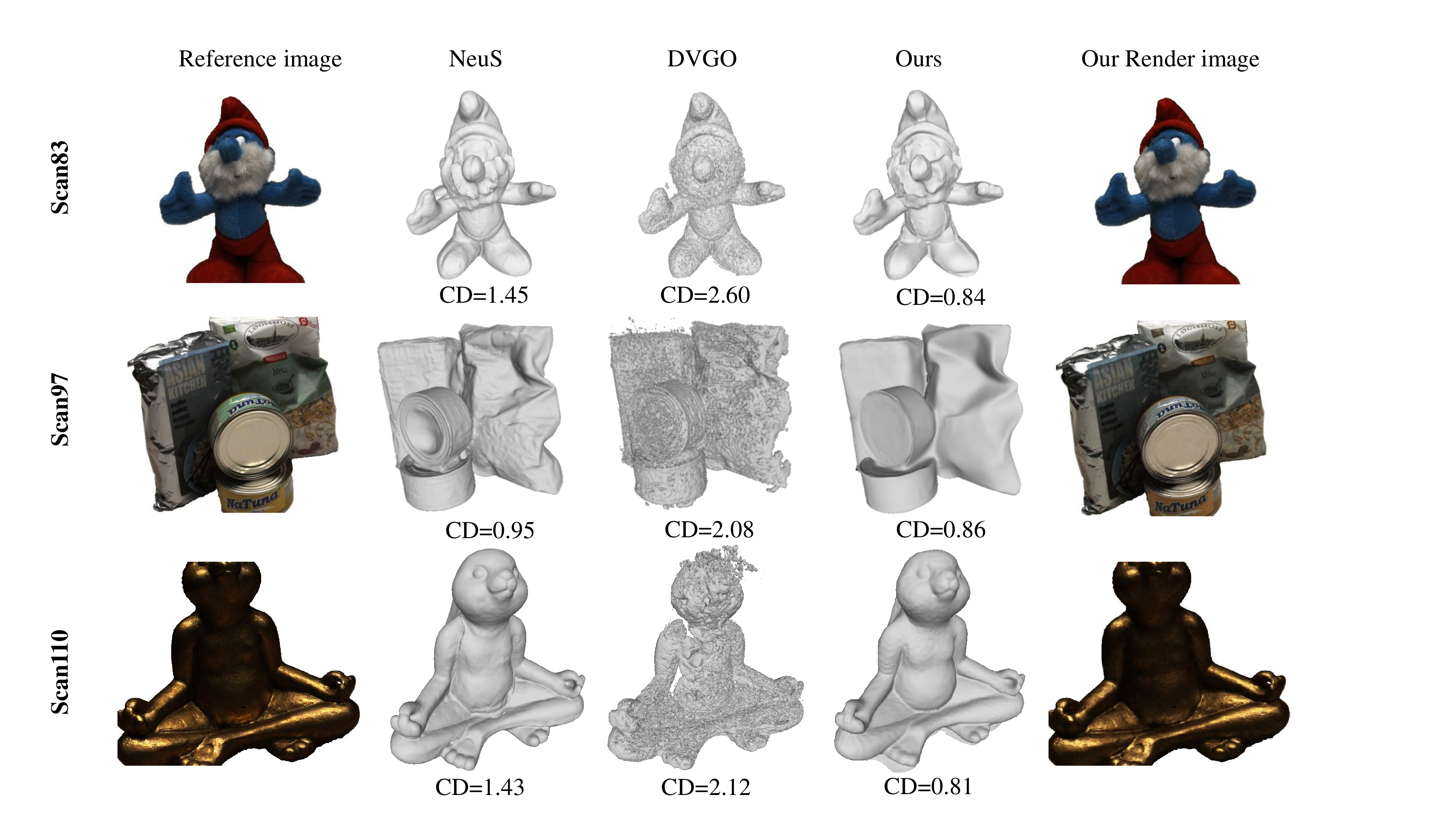}
  \end{center}
  \vspace{-0.4cm}
  \caption{{\bf Qualitative comparisons on DTU.}  In contrast to NeuS~\cite{Wang:2021:Neus} and DVGO~\cite{Sun:2022:Direct}, our approach is capable of recovering the substantial geometric details and synthesizing realistic novel view within a very short time.
  } 
  \label{fig:dturesult}
  \vspace{-0.5cm}
\end{figure*}

\subsection{Datasets and Baselines}
The DTU dataset~\cite{Jensen:2014:DTU} is employed as testbed for both quantitative and qualitative evaluations, which has been widely adopted by most of relevant studies. Additionally, the BlendedMVS~\cite{Yao:2020:blendedmvs}, NHR~\cite{Wu:2020:multi} and Tanks$\&$Temples~\cite{knapitsch:2017:tanks} datasets are utilized for visualization to demonstrate the scalability of our presented method. We compare the proposed approach with the following baseline methods for evaluation, including IDR~\cite{Yariv:2020:multiview}, NueS~\cite{Wang:2021:Neus}, NeRF~\cite{Mildenhall:2020:NeRF} and DVGO~\cite{Sun:2022:Direct}. IDR is a surface rendering-based method using implicit representation for 3D reconstruction. NueS is a volume rendering-based method with implicit representation.  NeRF achieves the high-performance on novel view synthesis method by volume rendering. DVGO accelerates NeRF through voxel-based explicit representation.

For the evaluation metrics on 3D reconstruction, we employ the Chamfer distance of the reconstructed model using the official evaluation code provided by the DTU dataset. As in~\cite{Wang:2021:Neus}, PSNR is used as the primary evaluation metric for novel view synthesis, where 90\% of the data is for training and the remaining 10\% for testing. To facilitate fair comparison, all methods are presented under the condition of clean background.

\subsection{Evaluation Results on DTU}

\begin{table}
	\centering
	\resizebox{1.0\linewidth}{!}{
		\begin{threeparttable}
			\begin{tabular}{c||c||c|c|c||c||c}
                \hline
                \multicolumn{1}{c||}{}&\multicolumn{1}{c||}{Points}&\multicolumn{3}{c||}{Implicit Function}&\multicolumn{1}{c||}{Voxel}&\multicolumn{1}{c}{Mesh} \\ 
				\hline
				ScanID & \thead{COLMAP\\\cite{Schoenberger:2016:mvs}} & \thead{IDR\\\cite{Yariv:2020:multiview}} & \thead{NeuS\\\cite{Wang:2021:Neus}} &  \thead{NeRF\\\cite{Mildenhall:2020:NeRF}} & \thead{DVGO\\\cite{Sun:2022:Direct}}  & Ours \\
				
				\hline
				% ScanID& COLM & IDR  & NeuS & NeRF & DVGO & NeuS & Ours \\
				\hline
				scan24  & \underline{0.81} & 1.62 & 0.83 & 1.83 & 1.83 & {\bf 0.65}  \\
				scan37  & 2.05 & 1.87 & {\bf 0.98} & 2.39 & 1.74 & \underline{1.48}  \\
				scan40  & 0.73 & 0.63 & {\bf 0.56} & 1.79 & 1.70 & \underline{0.57}  \\
				scan55  & 1.22 & 0.48 & {\bf 0.37} & 0.66 & 1.53 & \underline{0.40}  \\
				scan63  & 1.79 & {\bf 1.04} & \underline{1.13} & 1.79 & 1.91 & 1.48  \\
				scan65  & 1.58 & 0.79 & {\bf 0.59} & 1.44 & 1.91 & \underline{0.77}  \\
				scan69  & 1.02 & 0.77 & \underline{0.60} & 1.50 & 1.77 & {\bf 0.56}  \\
				scan83  & 3.05 & 1.33 & 1.45 & \underline{1.20} & 2.60 & {\bf 0.86}  \\
				scan97  & 1.40 & 1.16 & \underline{0.95} & 1.96 & 2.08 & {\bf 0.84}  \\
				scan105 & 2.05 & {\bf 0.76} & \underline{0.78} & 1.27 & 1.79 & 0.94  \\
				scan106 & 1.00 & \underline{0.67} & {\bf 0.52} & 1.44 & 1.76 & 0.72  \\
				scan110 & 1.32 & \underline{0.90} & 1.43 & 2.61 & 2.12 & {\bf 0.81}  \\
				scan114 & 0.49 & \underline{0.42} & {\bf 0.36} & 1.04 & 1.60 & 0.52  \\
				scan118 & 0.78 & 0.51 & {\bf 0.45} & 1.13 & 1.80 & \underline{0.49}  \\
				scan122 & 1.17 & \underline{0.53} & {\bf 0.49} & 0.99 & 1.58 & 0.54  \\
				\hline
				mean $\downarrow$    & 1.36 & \underline{0.90} & {\bf 0.77} & 1.85 & 1.54 & {\bf 0.77}  \\
				\hline
                Time[min] $\downarrow$  & 80 & 390 & 780 & 750 & {\bf 25}  & \underline{30} \\
                \hline
			\end{tabular}
			
		\end{threeparttable}
	}
    \vspace{-0.2cm}
	\caption{ {\bf Comparisons on reconstruction (Chamfer Distance) and training time on DTU.} Compared to NeuS, our method obtains $20\times$ speedup while achieving the better reconstruction results. Our method significantly outperforms DVGO with similar processing time. }
	\label{tbl:dtu_chamfer}
    \vspace{-0.2cm}
	%\end{wraptable}
\end{table}

\begin{table}
	\centering
	\resizebox{1.0\linewidth}{!}{
		\begin{threeparttable}
			\begin{tabular}{c||c|c||c|c|c}

                \hline
				ScanID & \thead{IDR\cite{Yariv:2020:multiview}} & \thead{NeuS\cite{Wang:2021:Neus}} &  \thead{NeRF\cite{Mildenhall:2020:NeRF}} & \thead{DVGO\cite{Sun:2022:Direct}}  & Ours \\
				
				\hline
				% ScanID& COLM & IDR  & NeuS & NeRF & DVGO & NeuS & Ours \\
				\hline
				scan24  & 23.29 & 26.13 & 26.97 & {\bf 27.77} & \underline{26.98}  \\
				scan37  & 21.36 & 24.08 & \underline{25.99} & 25.96 & {\bf 26.01}  \\
				scan40  & 24.39 & 26.73 & 27.68 & \underline{27.75} & {\bf 28.30}  \\
				scan55  & 22.96 & 28.06 & \underline{29.39} & {\bf 30.42} & 26.58  \\
				scan63  & 23.22 & 28.69 & \underline{33.07} & {\bf 34.35} & 29.25  \\
				scan65  & 23.94 & 31.41 & 30.87 & \underline{31.18} & {\bf 32.01}  \\
				scan69  & 20.34 & 28.96 & 27.90 & 29.52 & {\bf 32.54}  \\
				scan83  & 21.87 & 31.56 & \underline{33.49} & {\bf 36.94} & 33.42  \\
				scan97  & 22.95 & 25.51 & 27.43 & \underline{27.67} & {\bf 27.93}  \\
				scan105 & 22.71 & 29.18 & \underline{31.68} & {\bf 32.85} & 30.65  \\
				scan106 & 22.71 & \underline{32.60} & 30.73 & {\bf 33.75} & 30.77  \\
				scan110 & 21.26 & 30.83 & 29.61 & \underline{33.10} & {\bf 33.96}  \\
				scan114 & 25.35 & 29.32 & \underline{29.37} & {\bf 30.18} & 28.47  \\
				scan118 & 23.54 & \underline{35.91} & 33.44 & {\bf 36.11} & 34.27  \\
				scan122 & 27.98 & \underline{35.49} & 33.41 & {\bf 36.99} & 35.12  \\
				\hline
				mean $\uparrow$   & 23.20 & 29.63 & 30.07 & {\bf 31.64} & \underline{30.42}  \\
                \hline
                Render Time $\downarrow$ & 30s & 60s & 30s & \underline{0.75s} & {\bf 0.10s} \\
				\hline
			\end{tabular}
			
		\end{threeparttable}
	}
    \vspace{-0.3cm}
	\caption{{\bf Comparisons on novel view synthesis (PSNR).} Our method achieves the best performance in multiple scenes with the fastest rendering speed.}
	\label{tbl:psnr}
    \vspace{-0.4cm}
	%\end{wraptable}
\end{table}

We evaluate the performance of 3D geometric and 2D novel view synthesis on 15 scenes from the DTU dataset, each of which consists of 49 images and spans different material ranges with varying specularities, posing challenges for classic multi-view reconstruction methods. \cref{tbl:dtu_chamfer} reports the quantitative results of surface reconstruction on the DTU dataset. The quantitative experimental results demonstrate that our method achieves better reconstruction quality under the same conditions. In terms of speed, our method obtains 10-fold acceleration comparing to NeRF, IDR, and NeuS, and outperforms DVGO on reconstruction quality with similar speed. 

\cref{tbl:psnr} shows the evaluation results of novel view synthesis on the same dataset. It can be observed that our method still has an advantage in synthesizing new views. This is because our proposed approach is able to faithfully reconstruct the underlying geometry for better novel view synthesis.

\subsection{More Qualitative Results}
\begin{figure*}[h]
  \begin{center}
     \includegraphics[trim={0.1cm 0.1cm 0.1cm 0.1cm},clip,width=1.0\linewidth]{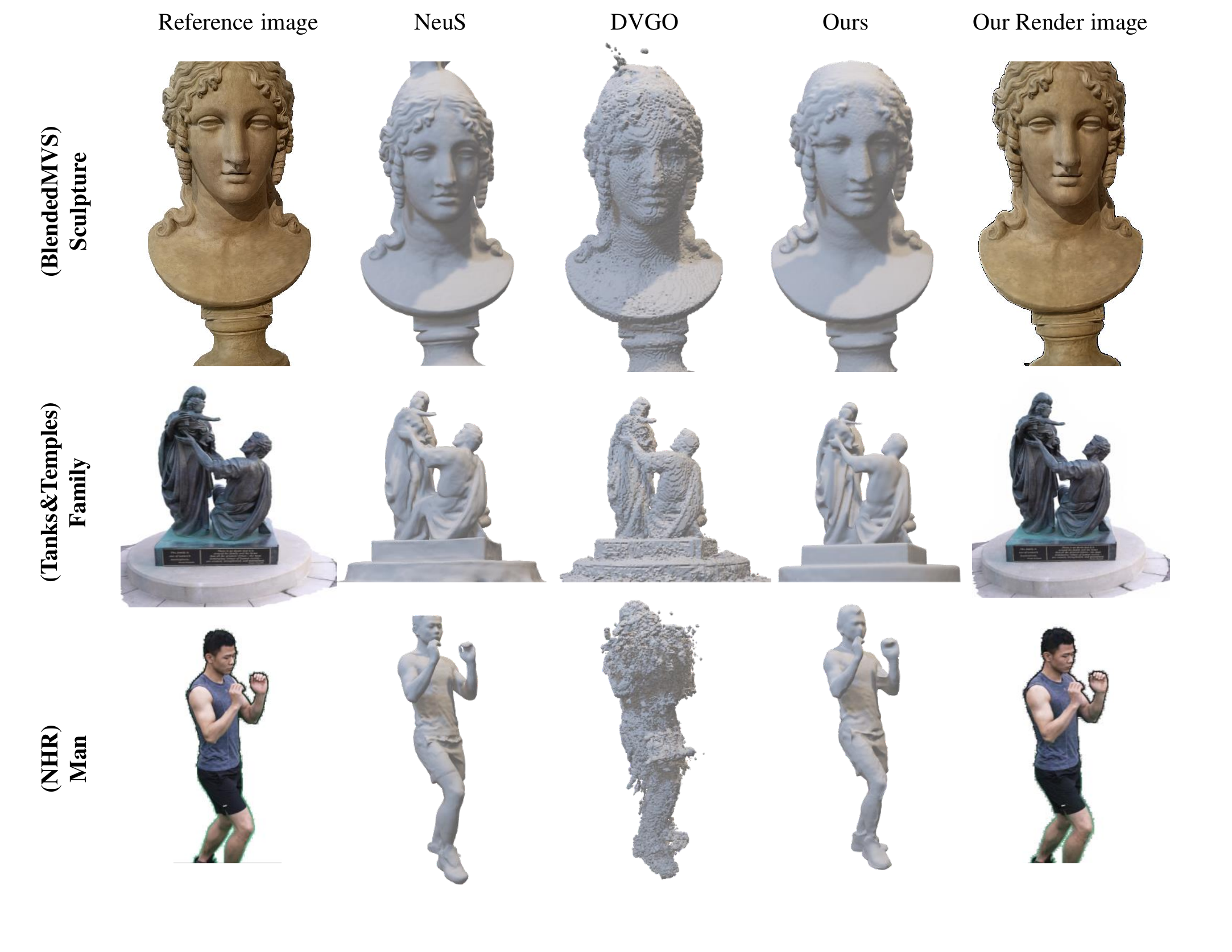}
  \end{center}
  \vspace{-1.0cm}
  \caption{{\bf Qualitative comparisons on challenging datasets.} We compare the reconstruction results of our method against the baseline methods on three challenging datasets, namely, the BlendedMVS~\cite{Yao:2020:blendedmvs}, Tanks$\&$Temples~\cite{knapitsch:2017:tanks}, and NHR~\cite{Wu:2020:multi}.
  } 
  \label{fig:other}
  \vspace{-0.2cm}
\end{figure*}

To evaluate the scalability and effectiveness of our presented method in outdoor large-scale scenes, we conducted experiments on the large-scale datasets, such as BlendedMVS, NHR, and Tanks$\&$Temples. Specifically, we use examples from the BlendedMVS low-resolution set with the size of $768\times576$. For the T$\&$T and NHR datasets, we employ images with the resolution of $1024 \times 1216$ and $540 \times 1024$, respectively. \cref{fig:other} shows the qualitative results.  Unlike the existing implicit surface or voxel-based methods, our proposed approach reconstructs the reasonable geometric structure and achieves the best novel view synthesis results in the shortest time.

\subsection{Ablation Study}

\begin{figure}[htb]
  \begin{center}
     \includegraphics[trim={0.1cm 0.1cm 0.1cm 0.1cm},clip,width=1.0\linewidth]{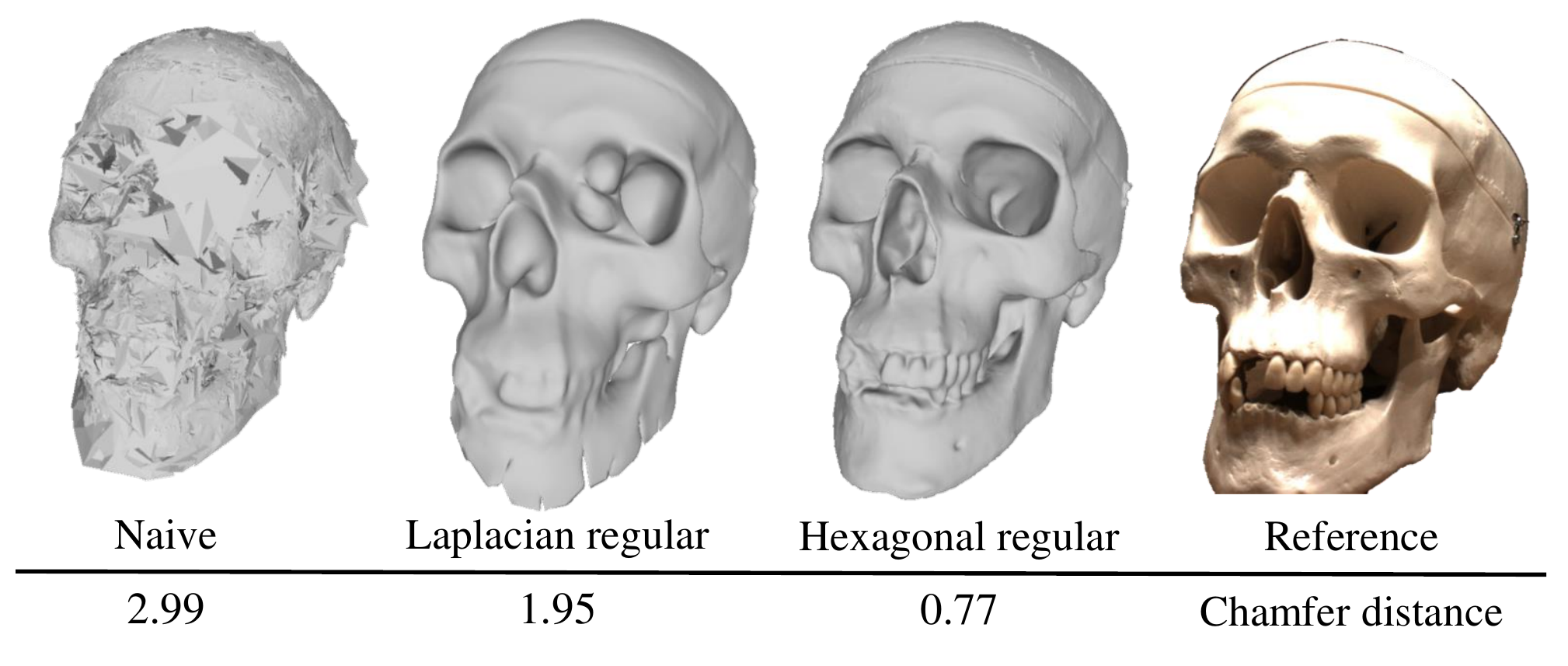}
  \end{center}
  \vspace{-0.4cm}
  \caption{ Ablation study on regularization.} 
  \label{fig:ablation_}
  % \vspace{-0.2cm}
\end{figure}

{\bf Mesh Regularization.} We compare the different mesh regularization techniques, including the naive method, the Laplacian regularization, and our presented hexagonal mesh regularization. \cref{fig:ablation_} shows the example results. It can be seen that the naive regularization of differentiable renderer produces a triangulated mesh with sever artifacts, in which the gradient steps pull on the silhouette without considering distortions or self-intersections. The Laplacian regularization is able to avoid this problem to some extent. However, the intersections of triangles still occur in the regions of high curvature.  Moreover, the strong constraints are prone to over-smoothing. Our presented method overcomes these issues of over-smoothing and self-intersections, which converges to a high-quality mesh.

{\bf Positional Encoding Level.} We conducted an ablation study on the level of positional encoding in our render network. Positional encoding has been shown to facilitate the learning of high-dimensional information by encoding the position in a specific way. Also, we aim to ensure the fast convergence of network. Thus, it is crucial to trade-off the level of positional encoding. Previous methods like UNISURF~\cite{Oechsle:2021:UNISURF} and VolSDF~\cite{Yariv:2021:volsdf} employ six levels of positional encoding, while NeRF~\cite{Mildenhall:2020:NeRF} uses ten levels. In contrast, we only use three levels of positional encoding. In \cref{tb:ab_level}, we give the quantitative evaluation results on the `scan83' from DTU dataset for different levels of positional encoding. The higher positional encoding levels improve the detail of rendered images, however, they incur some artifacts on surface.

\begin{table}[htbp]
  %\vspace{-0.1cm}
  \begin{center}
  \resizebox{0.75\linewidth}{!}{
  \begin{tabular}{l|ccc}
  \hline
  Method & Training time & PSNR & CD \\
  \hline
  level(d,v)=(3,0) & 32min & 30.98 & {\bf 0.86} \\
  \hline
  level(d,v)=(0,3) & 31min & 26.84 & 1.44 \\

  level(d,v)=(6,6) & 44min & 29.42 & 0.98 \\

  level(d,v)=(10,10) & 62min & {\bf 31.77} & 1.91 \\

  \hline
  \end{tabular}
  }
  \end{center}
  \vspace{-0.2cm}

  \caption{ Ablation study on positional encoding level. }% using scan83 of DTU.}
  \label{tb:ab_level}
 \vspace{-0.2cm}
\end{table}

\section{Conclusion}
This paper proposed a novel mesh-based neural rendering approach to fast multi-view reconstruction and novel view synthesis. Instead of using multiple points along the ray in volume rendering-based method, our approach only sampled at the intersection of ray and mesh, which enables to accurately disentangle geometry and appearance from input images.  A coarse-to-fine scheme was introduced to efficiently extract the initial mesh by space carving. Furthermore, the hexagonal mesh model was suggested to preserve surface regularity by constraining the second-order derivatives of vertices, where only a few level of positional encoding is sufficient for neural rendering. The promising experimental results demonstrate that our proposed approach is very effective for both multi-view reconstruction and novel view synthesis. In contrast to the implicit representation-based methods, a 10-fold acceleration in training was achieved.

{\small
\bibliographystyle{ieee_fullname}
\bibliography{egbib}
}

\end{document}